\documentclass[journal]{IEEEtran}

\usepackage{acronym}
\usepackage[utf8]{inputenc}
\usepackage{float}
\usepackage{tikz}
\usepackage{bm}
\usetikzlibrary{shapes,arrows}

\usepackage{arydshln}
\usepackage{cite}
\usepackage{graphicx}
\usepackage{subfig}
\usepackage{stfloats}
\usepackage{amsmath}
\usepackage{amsfonts}
\newcommand{\vc}[1]{\mathbf{\bm{#1}}} 
\renewcommand{\vec}[1]{\mathbf{\bm{#1}}} 

\pgfdeclarelayer{background}
\pgfdeclarelayer{foreground}
\pgfsetlayers{background,main,foreground}

\tikzstyle{eq}=[draw, fill=gray!20,
    text centered, rounded corners]
\tikzstyle{ann} = [above, text width=5em]
\tikzstyle{ukfeqs} = [eq, text width=10em, fill=gray!10,
    minimum height=12em, rounded corners, dashed]

\begin{document}

\title{\acs{UAV} attitude estimation using \acl{UKF} and \acs{TRIAD}}

\author{Hector~Garcia~de~Marina,~\IEEEmembership{Student,~IEEE,}
        Fernando~J.~Pereda,
        Jose~M.~Giron-Sierra,~\IEEEmembership{Member,~IEEE,}
        and Felipe~Espinosa,~\IEEEmembership{Member,~IEEE,}}

%
%
%

\newacro{AHRS}[AHRS]{Attitude Heading Reference System}
\newacro{DCM}[DCM]{Direct Cosine Matrix}
\newacro{EKF}[EKF]{Extended Kalman Filter}
\newacro{FAA}[FAA]{Federal Aviation Administration}
\newacro{GPS}[GPS]{Global Position System}
\newacro{GRV}[GRV]{Gaussian Random Variable}
\newacro{INS}[INS]{Inertial Navigation System}
\newacro{MEMS}[MEMS]{MicroElectroMechanical systems}
\newacro{TRIAD}[TRIAD]{Three Axis Attitude Determination}
\newacro{UAV}[UAV]{Unmanned Aerial Vehicle}
\newacro{UKF}[UKF]{Unscented Kalman Filter}
\newacro{IMU}[IMU]{Inertial Measurement Unit}
\newacro{UT}[UT]{Unscented Transformation}
\newacro{NED}[NED]{North East Down}

\maketitle

\begin{abstract}
A main problem in autonomous vehicles in general, and in \acp{UAV} in particular, is the
determination of the attitude angles. A novel method to estimate these angles using off-the-shelf
components is presented. This paper introduces an \ac{AHRS} based on the \ac{UKF} using the
\ac{TRIAD} algorithm as the observation model. The performance of the method is assessed through
simulations and compared to an \ac{AHRS} based on the \ac{EKF}. The paper presents field experiment
results using a real fixed-wing \ac{UAV}. The results show good real-time performance with low
computational cost in a microcontroller.
\end{abstract}

\begin{keywords}
\acf{AHRS}, \acf{UKF}, \acf{EKF}, \acf{UAV}, \acf{TRIAD}.
\end{keywords}

\IEEEpeerreviewmaketitle

\section{Introduction}

\acresetall

\PARstart{T}{here} is growing interest in autonomous vehicles. These vehicles are suitable for
mobile missions, specially in vigilance, monitoring and inspection scenarios
\cite{Maza2010,Cole2006}. Be it ground, marine or aerial, controlling an autonomous vehicle needs
some knowledge on its attitude angles \cite{Xia2011,Zheng2011,Zheng2011b}. These angles can be
measured in different ways, for instance, using a conventional \ac{INS}. Modern \ac{MEMS}
technologies are offering light and moderate cost solutions, denoted as \acp{IMU}, which are
appropriate for lightweight \acp{UAV}.

Our research group is involved in the development of \acp{UAV} relying on experimentation through a
spiral life cycle development based on prototypes. An on-board hardware and software have been
designed for these \acp{UAV}. The hardware includes an \ac{IMU} with three-axis accelerometers,
gyrometers and magnetometers; a \ac{GPS} receiver is also included. The system is light and small
enough to fly on a small fixed-wing \ac{UAV}.

The first flights, using manual control, have been used to gather signals from the sensors. We are
looking at closing autonomous control loops, which needs an accurate estimation of the three
attitude angles. The signals from the sensors have lots of vibrations due to the mechanical nature
of the system and noise with bias due to the sensors themselves and environmental effects. The
common solution in the literature is to use a kind of Kalman filter.

Conventional kinematic models of flying vehicles are highly non-linear; the filter should be able to
cope with these non-linearities. A typical Kalman filter for non-linear systems is the \ac{EKF}.
Using our experimental signals the performance of the \ac{EKF} was not completely satisfactory.
Sometimes estimation errors are too high and sometimes the filter may diverge. This is inappropriate
for control loops \cite{Cai2008}.

Searching for a better alternative we selected the \ac{UKF} as a potentially better solution. While
the \ac{EKF} is based on the linearization of the model through Jacobians or Hessians, the \ac{UKF}
uses the non-linear model directly; therefore the predictions should be more accurate.

The nine terms of the \ac{DCM} can be measured by using the \ac{TRIAD} algorithm and data from
on-board sensors: accelerometers and magnetometers. Using a quaternion formulation, which is a
conventional way to deal with the attitude of aerial vehicles \cite{Spinka2011}, the \ac{DCM} terms
can be easily handled. The quaternion approach is widely used in \acp{AHRS} because it avoids the
gimbal lock problem. With respect to numerical stability, quaternions are easier to propagate than
the angles themselves.

For systematic and non-risky work, a simulation environment has been developed. The core of the
simulation is the X-Plane 9 simulator which is certified by the \ac{FAA} for subsonic terrestrial
flight. For realism, perturbations in the form of high-frequency noise and sensor latencies, are
included to the data from the simulator.

For evaluating the proposed solution, a number of simulations have been run. These have been used to
both validate the solution and to compare it to the same algorithm using the \ac{EKF}. During these
simulations a maximum error of 1.0º is imposed on the pitch and roll attitude angles, and a maximum
error of 4.0º on the yaw attitude angle. Under those requirements, which are standard in the
industry \cite{nasa97}, the error tolerances in the different signals are obtained through the
analysis of a large number of simulation cases. The solution based on the \ac{UKF} shows better
performance.

\begin{figure}[htbp]
\centering
\includegraphics[width=0.4\textwidth]{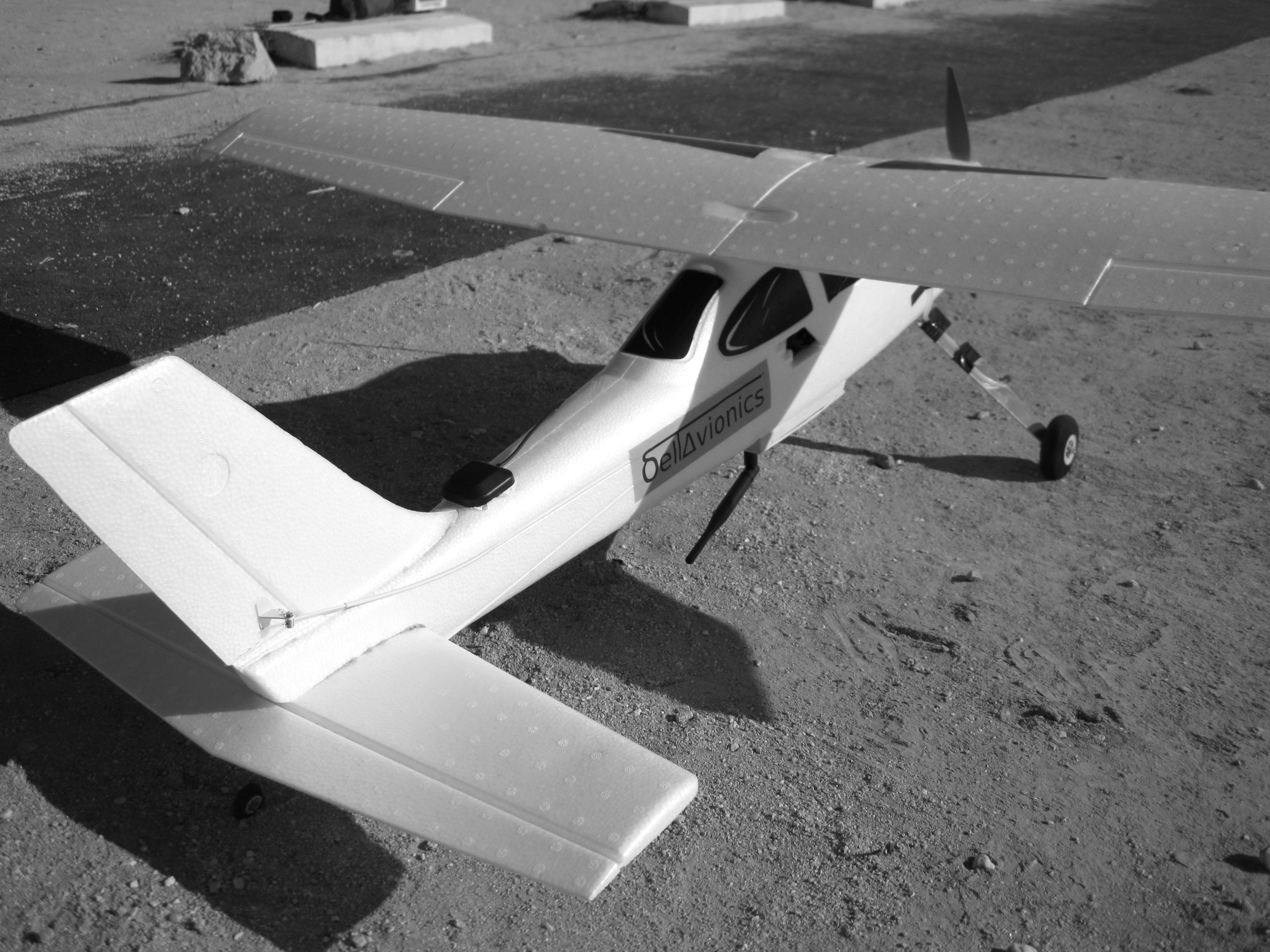}
\caption{\ac{UAV} developed by the authors.}
\label{fig: mentor}
\end{figure}

Finally, the \ac{UKF} based solution is tested on a real experiment using a fixed-wing \ac{UAV} of 2
meters of wingspan. This \ac{UAV} is shown in figure~\ref{fig: mentor}. An in-house autopilot is
used to test the estimation algorithm presented in this paper. The results of the solution are
compared to different independent systems with very good results.

Summarising, the first section of the paper presents some of the state of the art of \acp{AHRS}. The
following section is devoted to the formulation of the problem, including the kinematic model and
the \ac{TRIAD} algorithm. Then comes a section describing the whole solution using the \ac{UKF} and
\ac{TRIAD}. The fifth section deals with data obtained by simulation, and the comparison between the
\ac{UKF} and the \ac{EKF} is presented. Then, field experiments are presented to fully evaluate the
proposed \ac{AHRS}. In the final section of the paper, some conclusions are drawn, including some
guidelines for future work.

\section{Background and related work}

The \ac{UT} and the \ac{UKF} were introduced by Julier and Uhlmann \cite{Julier1997}, the algorithms
were further explained, with examples, by Wan and Merwe\cite{vanMerwe}. An extensive description of the \ac{UKF}
was given in \cite{Julier2004}.

Background on the \ac{EKF} algorithm is given in \cite{Jaz1970}. In \cite{Reif1999} it was shown
than the \ac{EKF} can lead to unbounded estimation errors for nonlinear systems similar to those
used in this paper.

Some modern techniques, such as particle filters \cite{Gustaf2002}, are expensive in computational
terms. Therefore they are unsuitable for microcontrollers unless the number of particles is reduced.
However, this can lead to worse performance than the \ac{EKF} \cite{Won2010}.

In a recent article \cite{Teixeira2010} a comparison of both the \ac{EKF} and \ac{UKF} is done for
the particular case of flight path reconstruction for a fixed-wing \ac{UAV}. This article includes an interesting discussion
of previous references that illustrate the improved performance of the \ac{UKF} over \ac{EKF}
estimating the attitude of an \ac{UAV}.

For real-time applications, it is important to reduce the computational cost. Part of the research
deals with different ways of alleviating this cost. For instance, in \cite{Marins2001} an attempt to
simplify the observation model is shown; however, it compromises computational cost because the
method involves the inversion of several large matrices every filter iteration. Another example is
\cite{Hale2004}, which proposes to reduce the number of vector state variables. Our proposed
solution has been implemented in a microcontroller for real-time estimation.

As highlighted by \cite{Qi2008} it is important to base the Kalman filter on an accurate model. In
our context, the models presented in the following section are well established in the literature
and correspond with experimental results.

An extensive review of navigation systems is \cite{Hasan2009}. It covers different algorithms
including Kalman filters and the \ac{TRIAD} algorithm. The \ac{TRIAD} algorithm was introduced by
Shuster and Oh in \cite{Shuster1981} to measure the \ac{DCM} in a spacecraft.

The use of \acp{IMU} based in \ac{MEMS} technology to estimate the attitude angles in the industry has been increasing in the recent years, like a fastening tool tracking system in \cite{Won2009}.

A well-known problem with gyrometers is bias. Therefore, different sensors have to be used to
correct these biases. For instance, in \cite{Du2010} they are corrected using three-axis
accelerometers. In \cite{Song2010} the use of eight accelerometers in a new configuration is
proposed for measuring angular velocities in small \acp{UAV}. Different
approaches rely on magnetometers \cite{Wu2008}, to be able to estimate yaw angle in helicopters, or
in \ac{GPS} \cite{Jang2007} to estimate the position as well the attitude in fixed-wing \acp{UAV}. Alternatively, other
papers propose not to use gyrometers at all in conventional aircrafts, but several \ac{GPS} receivers instead
\cite{Gebre2000}.

In this paper we propose an attitude estimator using the \ac{UKF} and the \ac{TRIAD} algorithm
involving gyrometers, magnetometers and accelerometers. The
validation of the algorithm is done by both simulations and field experiments. Experiments use an
on-board hardware with \ac{MEMS} sensors.

\section{Problem formulation}
Although each \ac{UAV} has its own coefficients and therefore its own dynamical model, it is
possible to determine Euler angles from a kinematic model, which is independent of the \ac{UAV}'s
coefficients. In this section we derive the mathematical formulation of the \ac{AHRS} problem in a
\ac{UAV} equipped with a three-axis gyrometer, three-axis accelerometer and three-axis magnetometer.

\subsection{\ac{AHRS} kinematic model}
\begin{figure}[htbp]
\centering
\includegraphics[width=0.25\textwidth]{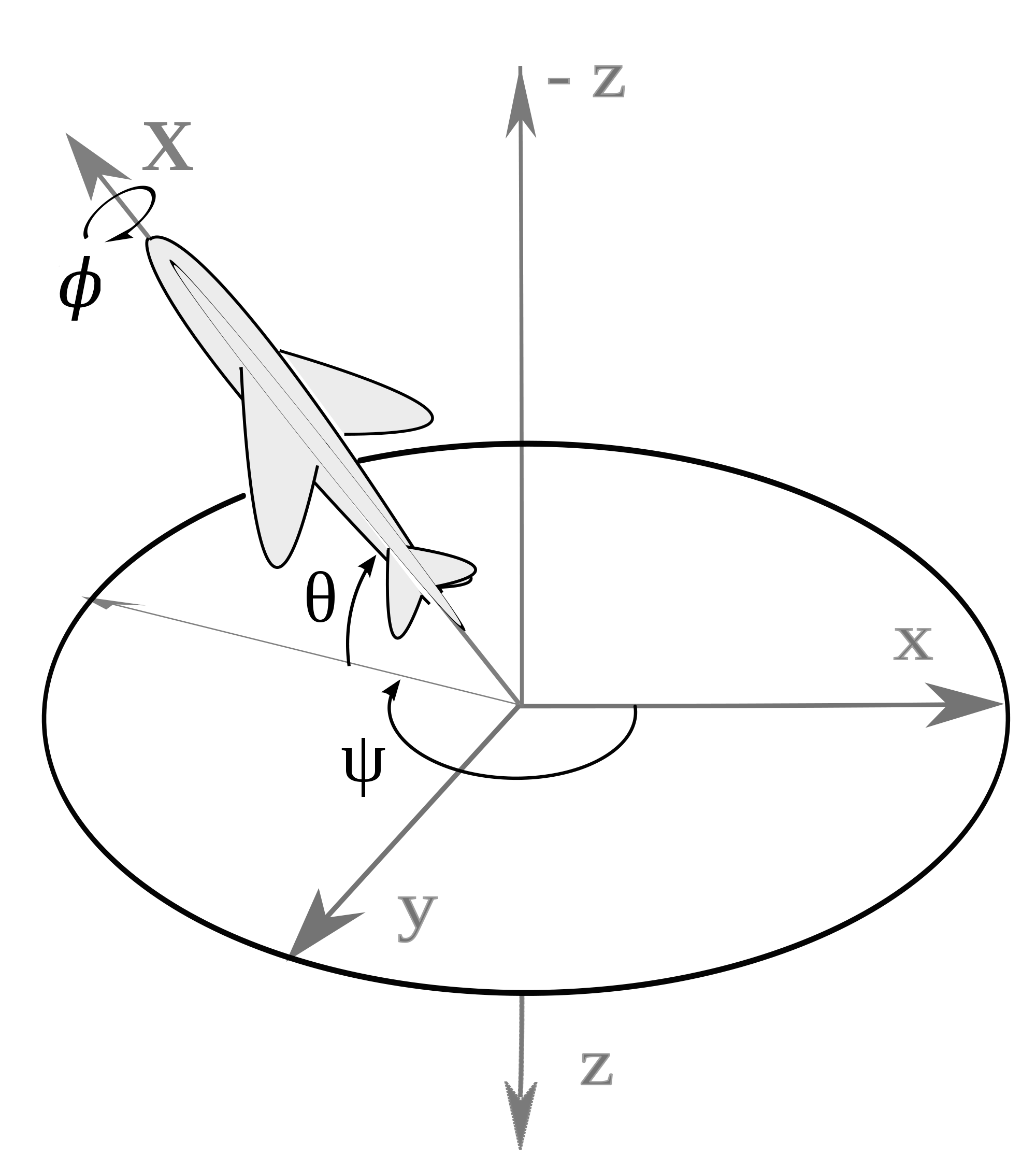}
\caption{Axes and coordinate definitions.}
\label{fig: coordinates}
\end{figure}

The Euler angles describe the aircraft body-axis orientation in north, east, and down coordinates.
That means in longitudinal, lateral and normal coordinates, with respect to the local tangent plane
to the Earth and true north. Here $\theta$ is the pitch angle, $\phi$ the roll angle and $\psi$ is
the yaw angle according to Figure \ref{fig: coordinates}. The angular velocity vector expressed in
body frame is $P$ for the roll rate, $Q$ is the pitch rate and $R$ is the yaw rate; and it is
related to the Earth frame by the transformation given by the kinematics equation \eqref{eq:
pqrbodyearth}.
\begin{equation}
\begin{bmatrix}
\dot{\phi} \\
\dot{\theta} \\
\dot{\psi}
\end{bmatrix}
=
\begin{bmatrix}
1 & \tan\theta \sin\phi & \tan\theta \cos\phi \\
0 & \cos\phi & -\sin\phi \\
0 & \frac{\sin\phi}{\cos\theta} & \frac{\cos\phi}{\cos\theta}
\end{bmatrix}
\begin{bmatrix}
P \\ Q \\ R
\end{bmatrix}
\label{eq: pqrbodyearth}
\end{equation}

Integrating equation \eqref{eq: pqrbodyearth} gives numerical instability and could be gimbal
locked. For this reason, a quaternion formulation to represent the attitude is preferred:
\begin{equation}
\begin{array}{l}
q = q_0 + q_1i + q_2j + q_3k \qquad \sum\limits_{i=0}^3 q_i = 1
\end{array}
\end{equation}

Where the quaternion norm is 1 and their components from Euler angles are:
\begin{align}
q_0 &= \cos\phi^\prime \cos\theta^\prime \cos\psi^\prime + \sin\phi^\prime
\sin\theta^\prime \sin\psi^\prime\\
q_1 &= \sin\phi^\prime \cos\theta^\prime \cos\psi^\prime - \cos\phi^\prime
\sin\theta^\prime \sin\psi^\prime\\
q_2 &= \cos\phi^\prime \sin\theta^\prime \cos\psi^\prime + \sin\phi^\prime
\cos\theta^\prime \sin\psi^\prime\\
q_3 &= \cos\phi^\prime \cos\theta^\prime \sin\psi^\prime - \sin\phi^\prime
\sin\theta^\prime \cos\psi^\prime
\end{align}
where $\phi^\prime = \phi/2$, $\theta^\prime = \theta/2$, and $\psi^\prime = \psi/2$.

The kinematics equation \eqref{eq: pqrbodyearth} can be rewritten in linear form using quaternion
components:
\begin{equation}
    \begin{bmatrix}
        \dot{q_0} \\
        \dot{q_1} \\
        \dot{q_2} \\
        \dot{q_3}
    \end{bmatrix} =
        \frac{1}{2}
        \begin{bmatrix}
            0 & -P & -Q & -R \\
            P &  0 &  R & -Q \\
            Q & -R &  0 &  P \\
            R &  Q & -P &  0
        \end{bmatrix}
        \begin{bmatrix}
            q_0 \\
            q_1 \\
            q_2 \\
            q_3
        \end{bmatrix}
\label{eq: pqr}
\end{equation}

Additionally, it is useful to formulate the \ac{DCM} using quaternion components and the Euler
angles from the \ac{DCM} terms:
\begin{align}
    &\text{DCM}\equiv A = \{c_{ij}\} = \nonumber \\
    & \begin{bmatrix}
        A_1                & 2(q_1q_2 + q_0q_3) & 2(q_1q_3 - q_0q_2) \\
        2(q_1q_2 - q_0q_3) & A_2                & 2(q_2q_3 + q_0q_1) \\
        2(q_1q_3 + q_0q_2) & 2(q_2q_3 - q_0q_1) & A_3
    \end{bmatrix} \label{eq: dcm_intro}
\end{align}
where $A_1 = q_0^2 + q_1^2 - q_2^2 - q_3^2$, $A_2 = q_0^2 - q_1^2 + q_2^2 - q_3^2$, and $A_3 = q_0^2
- q_1^2 - q_2^2 + q_3^2$. Then,
\begin{align}
    \theta &= -\arcsin(2(q_1q_3 - q_0q_2)) \\
    \phi   &= \operatorname{atan2}(2(q_2q_3 + q_0q_1), q_0^2 - q_1^2 - q_2^2 + q_3^2) \\
    \psi   &= \operatorname{atan2}(2(q_1q_2 + q_0q_3), q_0^2 + q_1^2 - q_2^2 - q_3^2)
\end{align}
where $\operatorname{atan2}$ is the four-quadrant version of the inverse tangent function, and
$\arcsin$ is the arcsine function.

\subsection{Gyros integration problem}
\label{subsec:gyros_integration}

The three-axis gyrometer measures the angular velocities, and for obtaining the Euler angles, the
gyros can be integrated using equation \eqref{eq: pqr}. However, even if we ignore the sensor noise,
the gyros usually have bias, making integration their error grow in every step.

Fortunately for a \ac{MEMS} gyrometer in normal conditions (not extremal temperature or pressure
variation), this bias can be assumed to be constant \cite{Hadri2009}, or very slow varying
throughout the \ac{UAV} mission. Therefore the bias for the gyrometer can be modeled as:
\begin{equation}
    \dot{\vc{b}} = 0 \quad \text{with} \quad \vc{b} = \begin{bmatrix} b_x b_y b_z \end{bmatrix}^T
    \label{eq: bias}
\end{equation}

Denoting the angular velocity vector $\vc{\omega} = (P,Q,R)$, if the gyros from the sensors are
$\vc{\omega}_s$, they can be corrected using the bias as:
\begin{align}
    \vc{\omega} = \vc{\omega}_s - \vc{b}
\end{align}

Another issue arises when the equation \eqref{eq: pqr} is integrated. The quaternion shall conserve
its norm equals to 1, otherwise the Euler angles are wrongly computed from the quaternion. Although
it can be normalized after every integration step, there is a better way to do it. The quaternion
norm can be kept solving the equation \eqref{eq: pqr} using the next integrating factor:
\begin{align}
    \exp\left(\int\limits_{t_0}^t\Omega \, \operatorname{d}t\right)
\end{align}
where
\begin{align}
\Omega = \frac{1}{2}
    \begin{bmatrix}
        0 & -P & -Q & -R \\
        P &  0 &  R & -Q \\
        Q & -R &  0 &  P \\
        R &  Q & -P &  0
    \end{bmatrix}
\end{align}

Assuming that the angular velocities remain constant during the interval $\operatorname{d}t$, we can
discretize the equation \eqref{eq: pqr} as follows\cite{Vaganay1993}:
\begin{align}
q(k+1) = \left(I \cos\frac{\|\Delta \vc{\omega}\|}{2} + \sin\frac{\|\Delta
\vc{\omega}\|}{2}{\|\Delta \vc{\omega}\|}\Omega\right) q(k)
\label{eq: pqrquat}
\end{align}
where $\|\Delta \vc{\omega}\| = \frac{1}{2} \sqrt{(P\Delta t)^2 + (Q\Delta t)^2 + (R\Delta
t)^2}$ and $I$ is the identity matrix.

\subsection{The \ac{TRIAD} algorithm}
\label{subsec: triad}

The \acl{TRIAD} algorithm was introduced by Shuster and Oh\cite{Shuster1981} to determine the
attitude in a Spacecraft from a set of vector measurements. \ac{TRIAD} is a deterministic method to
compute the \ac{DCM}. Given the reference unit vectors $\vec{V_1}$ and $\vec{V_2}$ and the
corresponding observation unit vectors $\vec{W_1}$ and $\vec{W_2}$, the \ac{DCM} satisfies:
\begin{align}
    A\vec{V_1} = \vec{W_1} \qquad A\vec{V_2} = \vec{W_2} \label{eq: triad_reference_pairs}
\end{align}

The \ac{TRIAD} algorithm determines the \ac{DCM} using the following expression
\begin{align}
    A   &= M_o M_r^T \\
    M_o &= \left(\vec{o_1} | \vec{o_2} | \vec{o_3}\right) \\
    M_r &= \left(\vec{r_1} | \vec{r_2} | \vec{r_3}\right)
\end{align}
where the observation column vectors $\vec{o_i}$ and the reference column vectors $\vec{r_i}$ are given by
\begin{align}
\vec{o_1} &= \vec{W_1} \\
\vec{o_2} &= \left(\vec{W_1}\times \vec{W_2}\right) / |\vec{W_1}\times \vec{W_2}| \\
\vec{o_3} &= \left(\vec{W_1}\times \left(\vec{W_1}\times \vec{W_2}\right) \right) / |\vec{W_1}\times
\vec{W_2}| \\
\vec{r_1} &= \vec{V_1} \\
\vec{r_2} &= \left(\vec{V_1}\times \vec{V_2}\right) / |\vec{V_1}\times \vec{V_2}| \\
\vec{r_3} &= \left(\vec{V_1}\times \left(\vec{V_1}\times \vec{V_2}\right) \right) / |\vec{V_1}\times
\vec{V_2}|
\end{align}

Notice that the pair $(\vec{W_1},\vec{V_1})$ has more influence on $A$ than $(\vec{W_2},\vec{V_2})$,
this is because part of the information contained in the second pair is discarded. Therefore it is
convenient to assign $(\vec{W_1},\vec{V_1})$ to the pair of greater accuracy, which depends on the
flight circumstances. Section \ref{subsec: criteria} describes criteria to assign these pairs.

\section{\acl{UKF} design}

The \ac{UKF} \cite{Julier1997} is an alternative to the \ac{EKF}, providing superior performance at
similar order of computational cost. Also during the \ac{UKF} implementation process, there are not
Jacobians, Hessians or other derivatives involved. In this paper, the \ac{UKF} is used both for
state estimation (quaternion components) and for parameter estimation (gyrometer's biases).

The estimation algorithm is implemented as a two-step \emph{propagator/corrector} filter. It is
desirable to run each step as many times as possible, however, the frequencies at which they will be
run will be limited by different factors. For the propagation step, the limiting factor is the
computation time; we chose a frequency of 100~Hz. The correction step is limited, in our case, by
the \ac{GPS} sample rate, that is, 1~Hz.

\begin{figure}[htbp]
\centering
\includegraphics[width=0.50\textwidth]{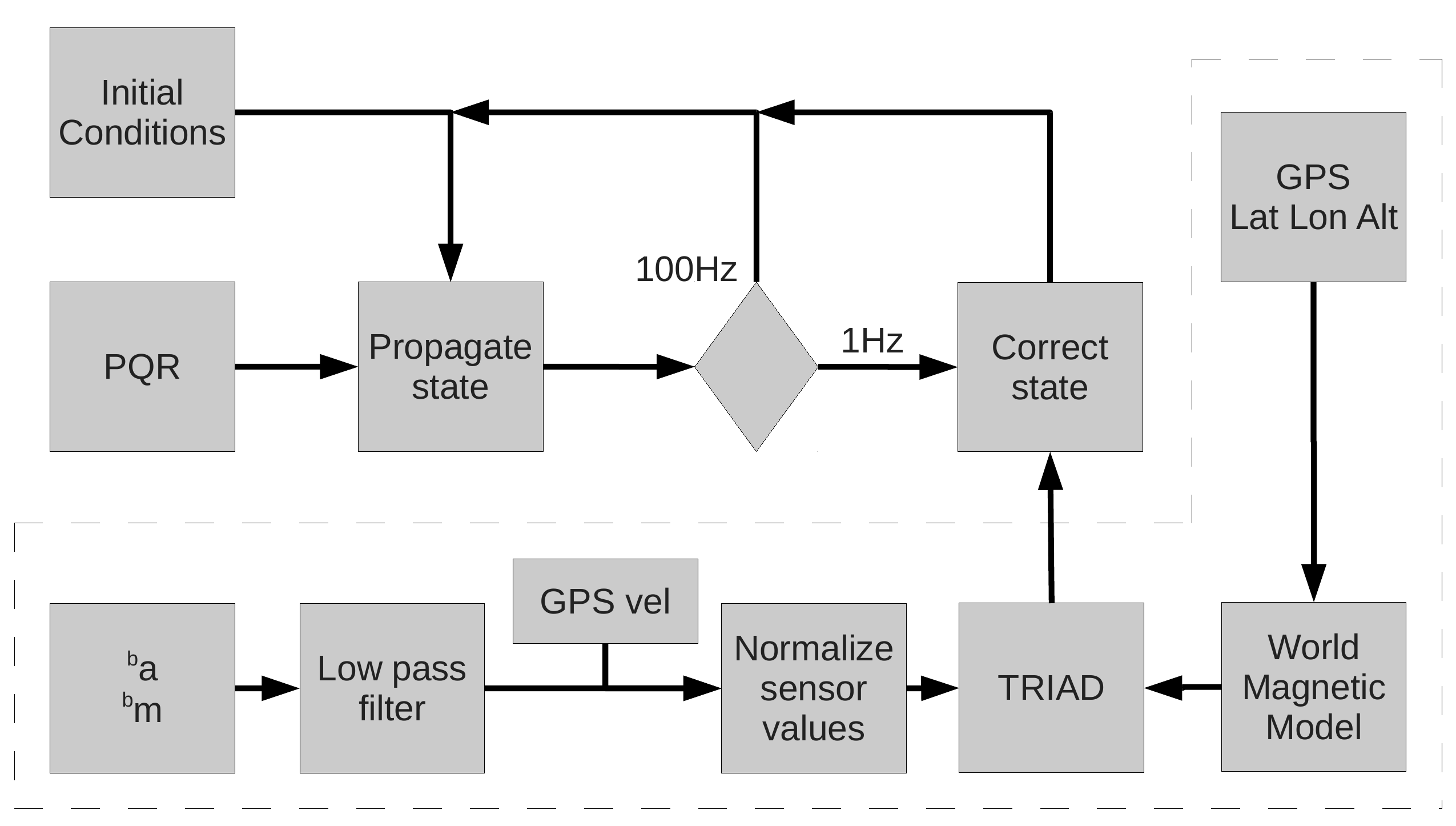}
\caption{Algorithm block diagram.}
\label{fig: filtro}
\end{figure}

The algorithm is described in the diagram shown in Figure~\ref{fig: filtro}. The diagram depicts the
two main loops, one at 100~Hz and the other at 1~Hz, these loops correspond to the propagation and
correction steps respectively. It can be noted how the \ac{TRIAD} information feeds the correction
step.

The area surrounded by dashed lines contains the elements involved in the \ac{TRIAD} calculations.
The leftmost block represents an access to the sensors to measure $^ba$ and $^bm$ which correspond
to the accelerations and magnetic field measurements in the body reference frame respectively. The
signals are then filtered. The \ac{GPS} velocity is used to subtract the centrifugal accelerations
as explained in subsection~B; which also covers the description of the \ac{TRIAD} block. On the
right side of the diagram the block denoted as \emph{World Magnetic Model} uses an harmonic
spherical model to obtain the magnetic field vector at the position of the \ac{UAV}. This vector is
used as one of the references in the \ac{TRIAD} algorithm.

This section describes the design of the filter under the considerations given above. Subsection~A
gives the equations involved in the propagation loop whilst Subsection~B deals with those of the
correction loop.

\subsection{Propagation equations and process model}

This subsection describes the propagation loop. The state vector is:
\begin{align}
x(k) = \begin{bmatrix}q_0(k) & q_1(k) & q_2(k) & q_3(k) & b_x(k)
& b_y(k) & b_z(k)\end{bmatrix}^T
\end{align}

Where $q_i$ are the quaternion components and $b_j$ are the gyrometers' biases. These are assumed to
be \acp{GRV}. Their process model is given by equations \eqref{eq: pqrquat} and \eqref{eq: bias},
respectively.

The basis of the \ac{UKF} is the \ac{UT}; which is a method for calculating the
statistics of a random variable which undergoes a nonlinear transformation \cite{Julier1997}. In our
case, we assume that the noise of the \ac{UAV} sensors is additive (zero mean). Under this
assumption, the equations to estimate the Euler angles and the gyrometer's biases are the ones
described in this subsection.

The computation algorithm begins with the initial conditions:
\begin{align}
    \hat{x}_0 &= E[x_0] \\
    P_0 &= E[(x_0 - \hat{x}_0)(x_0 - \hat{x}_0)^T]
\end{align}

At the start of every iteration we calculate the \emph{sigma points} \cite{Julier1997} $\chi_{k-1}$
of the quaternion components and gyro's biases.
\begin{align}
    \chi_{k-1} =
        \begin{bmatrix}
            \hat{x}_{k-1} &
            \hat{x}_{k-1}+\gamma\sqrt{P_{k-1}} &
            \hat{x}_{k-1}-\gamma\sqrt{P_{k-1}}
        \end{bmatrix}
\end{align}

Where $\gamma = \sqrt{L+\lambda}$, $\lambda$ is a composite scaling parameter and $L$ is the state
vector dimension. $\sqrt{P_k}$ can be computed with the lower triangular Cholesky factorization.

The next step is to evaluate the model on the sigma points:
\begin{align}
    \chi_{k|k-1}^* = F[\chi_{k-1},u_{k-1}]
\end{align}

Where $F$ is the process model as given by equations \eqref{eq: pqrquat} and \eqref{eq: bias}.
$u_{k-1}$ is the angular velocity vector $\vec{\omega}$. The a priori state estimation is
approximated using a weighted sample mean
\begin{align}
    \hat{x}_{k}^- = \sum\limits_{i=0}^{2L}W_i^{(m)}\chi_{i,k|k-1}^*
\end{align}
where
\begin{align}
    W_0^{(m)} &= \lambda/(L + \lambda) \\
    W_i^{(m)} &= 1/\left(2(L + \lambda)\right)
\end{align}
and the covariance matrix is
\begin{align}
    P_k^- = \sum\limits_{i=0}^{2L}W_i^{(c)}[\chi_{i,k|k-1}^* - \hat{x}_k^-]
        [\chi_{i,k|k-1}^* - \hat{x}_k^-]^T + Q_{\text{\tiny noise}}
\end{align}
where
\begin{align}
    W_0^{(c)} &= \lambda/(L + \lambda) + (1 - \alpha^2 + \beta) \\
    W_i^{(c)} &= 1/\left(2(L + \lambda)\right)
\end{align}

The constant $\alpha$ determines the spread of the sigma points around $\hat{x}$ and is usually set
to a small positive value (for instance, $1 \leq \alpha \leq 1e^{-4}$). The constant $\beta$ can be
used to incorporate prior knowledge of the distribution of the state vector: for Gaussian
distributions, $\beta = 2$ is optimal \cite{vanMerwe}. $Q_{\text{\tiny noise}} \in \mathbb{R}^{7x7}$
is the process noise covariance:
\begin{align}
    Q_{\text{\tiny noise}} =
        \begin{bmatrix}
            \begin{array}{c;{1pt/1pt}c}
                Q_q & 0 \\ \cdashline{1-2}[1pt/1pt]
                0 & Q_b
            \end{array}
        \end{bmatrix}
\end{align}
where $Q_q \in \mathbb{R}^{4x4}$ is the noise covariance associated to the quaternion components and
$Q_b \in \mathbb{R}^{3x3}$ is the noise covariance associated to the gyrometer biases.

The process that relates the quaternion components and angular velocities has a continuous-time
analytical solution, as was shown in Subsection~\ref{subsec:gyros_integration}. However, the
discrete-time equation~\eqref{eq: pqrquat} assumes that the angular velocities remain constant
during the discretization period. Hence, $Q_q$ should be close, but different to zero. For the
simulation and experimental results shown in this paper:
\begin{align}
    Q_q = 1\times10^{-6} \cdot I_{4x4}
\end{align}
where $I_{4x4}$ is the 4x4 identity matrix.

The process noise covariance associated to the gyrometer biases $Q_b$ is the 3x3 zero matrix. The
rationale for this is the same that explains equation~\eqref{eq: bias} in
Subsection~\ref{subsec:gyros_integration}.

\subsection{Correction equations and observation model}
\label{subsec: criteria}

Even though the \ac{TRIAD} algorithm gives the nine terms of the \ac{DCM} (see equation~\eqref{eq:
dcm_intro}), only four of them are needed to calculate the Euler angles. Hence, the observation
function has been designed to measure these four terms.

To determine the pitch and roll angles, the terms are the X and Y components of the Z earth vector
expressed in the body frame.
\begin{align}
    {^bZ_{Ex}} &\equiv c_{13} = 2(q_1q_3 - q_0q_2) \\
    {^bZ_{Ey}} &\equiv c_{23} = 2(q_2q_3 + q_0q_1)
\end{align}

For the yaw angle, the terms are the X and Y components of the X body vector expressed in the Earth
reference frame.
\begin{align}
    {^EX_{bx}} &\equiv c_{11} = (q_0^2 + q_1^2 - q_2^2 - q_3^2) \\
    {^EX_{by}} &\equiv c_{12} = 2(q_1q_2 + q_0q_3)
\end{align}

Therefore, the observation model is given by equation~\eqref{eq: observation_model}.
\begin{align}
    H(x_k) =
        \begin{bmatrix}
            c_{13} & c_{23} & c_{11} & c_{12}
        \end{bmatrix}^T
    \label{eq: observation_model}
\end{align}

According to the \ac{TRIAD} algorithm two vector pairs are needed to compute the terms of the
\ac{DCM}. Each pair consists of a measure or observation and a reference vector (equation~\eqref{eq:
triad_reference_pairs}). In our case, these pairs are the magnetic field, and the acceleration of
the \ac{UAV}.

The magnetic observation vector is the field measured at body frame. The magnetic reference vector
is the Earth's magnetic field using an harmonic spherical model; the geographical coordinates for
this model are provided by an on-board GPS receiver.

For the acceleration pair, the observation is given by the on-board \ac{IMU}. It should be noted
that the measurement given by the accelerometers includes: linear acceleration, Coriolis
acceleration, centripetal acceleration and gravity. Coriolis acceleration is assumed to be
negligible, in our experiments it is of the order of $10^{-4}$ g. The reference is the Earth's gravity
$\hat{\vec{g}} = \begin{bmatrix}0 & 0 & 1\end{bmatrix}$, normalized in \ac{NED} coordinates. The
on-board accelerometers are affected by mechanical vibrations and environmental perturbations,
therefore it is convenient to use a low-pass filter to reduce these effects. Also, the centrifugal
contribution of the acceleration has to be subtracted:
\begin{align}
    {^Ea_N} &= \dot{U} + QW - RV \\
    {^Ea_E} &= \dot{V} + RU - PW \\
    {^Ea_D} &= \dot{W} + PV - RU
    \label{eq: centrifugal}
\end{align}

Where the speed $^b\vec{V} = \begin{bmatrix}U & V & W\end{bmatrix}$ is expressed in the body frame
of reference. In a fixed-wing \ac{UAV}, during a non-acrobatic flight, both $V$ and $W$ are
negligible. Therefore, the \ac{GPS} speed measurement gives $U$, which is the dominant component of
$^b\vec{V}$.

As stated in section~\ref{subsec: triad} it is convenient to assign $(\vec{W_1},\vec{V_1})$ to the
pair of greater accuracy. During a flight, there are times when accelerometers offer more accuracy,
whereas during others, magnetometers are more reliable. In our case, we used the following criteria.
In order of priority:

\begin{itemize}
    \item If $0.9\hat{\vec{g}} \leq |{^b\vec{a}}| \leq 1.1\hat{\vec{g}}$ then it can be assumed that
    this is a stationary flight, hence the acceleration is the pair of greater accuracy. In this
    case, $\vec{V_1} = \hat{\vec{g}}$ and $\vec{W_1} = {^b\vec{a}}$.
    \item Else, if $0.7\hat{\vec{g}} < |{^b\vec{a}}| < 0.9\hat{\vec{g}} \quad \text{or} \quad
    1.1\hat{\vec{g}} < |{^b\vec{a}}| < 1.3\hat{\vec{g}}$ then it can be assumed that the plane is
    doing a coordinated turn. In this case, magnetometers offer greater accuracy.  Therefore,
    $\vec{V_1} = {^E\vec{m}}$ and $\vec{W_1} = {^b\vec{m}}$.
    \item Else, if $|{^b\vec{m}}| > 1.2 |{^E\vec{m}}| \quad \text{or} \quad |^b\vec{m}| < 0.8
    |{^E\vec{m}}|$ then the magnetic measurements are not reliable. The correction step is skipped.
    \item Else, if $|{^b\vec{a}}| > 1.3\hat{\vec{g}} \quad \text{or} \quad |{^b\vec{a}}| <
    0.7\hat{\vec{g}}$ then it is assumed that the current state is acrobatic and neither the
    magnetic measurements nor the acceleration measurements are reliable enough. The correction step
    is skipped.
\end{itemize}

Where ${^E\vec{m}}$ is the Earth's magnetic field vector and $\hat{\vec{g}}$ the Earth's gravity
acceleration. Notice how in the last two cases, the correction step is skipped. The following only
applies in the first two cases.

The \ac{UKF} begins the correction step by redrawing the sigma points. This is done to incorporate
the effect of the additive noise \cite{vanMerwe}.
\begin{align}
\chi_{k|k-1} =
    \begin{bmatrix}
        \hat{x}_k^- &
        \hat{x}_k^-+\gamma\sqrt{P_k^-} &
        \hat{x}_k^--\gamma\sqrt{P_k^-}
    \end{bmatrix}
\end{align}

Then the unscented transformation of the observations is computed.
\begin{align}
    \mathcal{Y}_{k|k-1} = H[\chi_{k|k-1}] \\
    \hat{y}_{k}^- = \sum\limits_{i=0}^{2L}W_i^{(m)}\mathcal{Y}_{i,k|k-1}
\end{align}

The measure and cross-covariance matrices are
\begin{align}
P_{\bar{y_k}\bar{y_k}} &= \sum\limits_{i=0}^{2L}W_i^{(c)}[\mathcal{Y}_{i,k|k-1} - \hat{y}_k^-]
[\mathcal{Y}_{i,k|k-1} - \hat{y}_k^-]^T + R_{\text{\tiny noise}} \\
P_{x_ky_k} &= \sum\limits_{i=0}^{2L}W_i^{(c)}[\chi_{i,k|k-1} - \hat{x}_k^-]
[\mathcal{Y}_{i,k|k-1} - \hat{y}_k^-]^T
\end{align}
where $R_{\text{\tiny noise}}$ is the measurement noise covariance. The value of $R_{\text{\tiny
noise}}$ can be derived from the nominal values of the errors of the sensors involved. This
derivation is described thoroughly in \cite{Shuster1981}.

Now the Kalman gain is computed:
\begin{align*}
    \mathcal{K}_k = P_{x_ky_k} P_{x_ky_k}^{-1}
\end{align*}

Finally, the correction equations are \eqref{eq: correct_mean} and \eqref{eq: correct_cov}:
\begin{align}
    \hat{x}_k &= \hat{x}_k^- + \mathcal{K}_k(y_k - \hat{y}_k^-) \label{eq: correct_mean} \\
    P_k &= P_k^- - \mathcal{K}_k P_{\bar{y_k}\bar{y_k}} \mathcal{K}_k^T \label{eq: correct_cov}
\end{align}

\section{Simulations results}

Since real experiments might imply crashes, some previous simulations are in order. For this, we
developed a simulation framework, which consists in three different parts. The core of the
simulation is the X-Plane 9 software. The other two parts are: the plug-in code for X-Plane 9 and
the model of the sensors. The idea is to integrate a six-degree-of-freedom aerodynamic model,
provided by X-Plane with a realistic model of the sensors we are using. The plug-in code is just the
glue between them.

X-Plane 9 includes different aircraft models. Its default radio control model is very similar to
our \ac{UAV}, so no modifications to it are needed. This is the model used in the simulations of
this section.

The purpose of the simulations is to study the effects of sensor noise, bias and latencies.
Therefore, the model of the sensors focuses on these aspects:

\begin{itemize}
    \item \ac{GPS} signal is delayed 1 second.
    \item Gyrometers signal are biased and corrupted with white noise.
    \item Accelerometers are biased and corrupted with colored noise, focusing in high frequencies.
    \item Magnetometers are biased and corrupted with white noise.
\end{itemize}

\begin{figure}[htbp]
\includegraphics[width=1\linewidth]{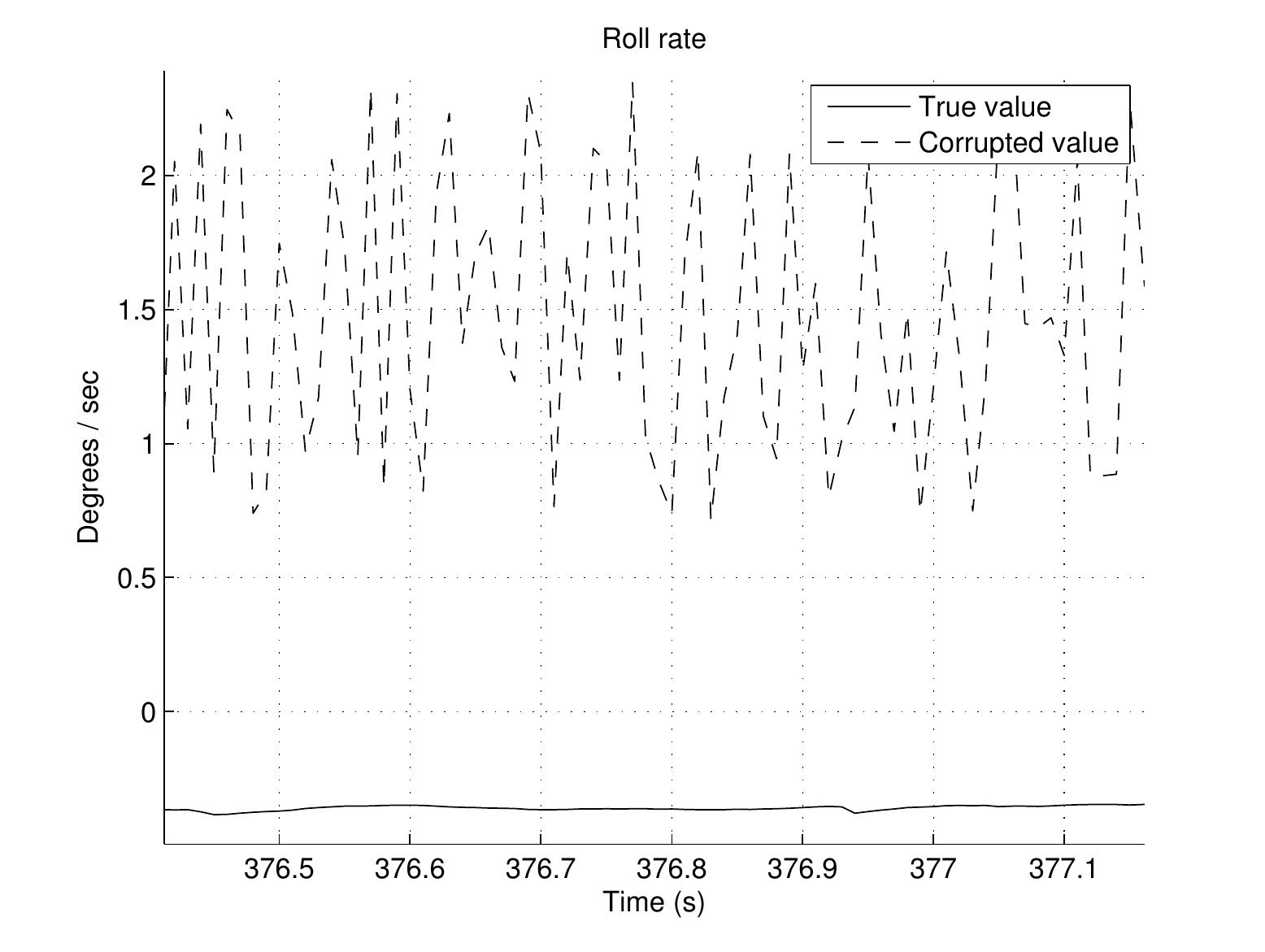}
\caption{True and corrupted roll rate used for simulations.}
\label{fig: p_corrupted}
\end{figure}

Figure~\ref{fig: p_corrupted} shows an example of how bias and noise are added to a variable. Notice
how both the magnitude of the noise and the bias are kept constant throughout the simulation. In
particular, the example shows roll rate measurements as simulated.

A first target of the simulations is to study the tolerances of both estimators to bias and noise
magnitude. According to standard procedures, a maximum error of 1.0º in the estimation of pitch and
roll angle, and 4.0º for yaw angle are imposed. It is assumed that with this error it is possible to
do closed-loop control. This is covered in the first part of this section. The second part of this
section assesses the performance of the \ac{UKF} using real values for the biases and noise
magnitudes of our sensors.

\subsection{Error tolerances and comparison of estimators}

A Monte Carlo analysis of the tolerances was made, supported by a batch of simulation experiments.
Each of the experiments specifies different values of biases and noise magnitudes, which were drawn
from a Gaussian distribution.

Two sets of results were obtained, one using the \ac{UKF} and the other using the \ac{EKF}. These
sets are shown in Tables \ref{tab: ukfvalues} and \ref{tab: ekfvalues} respectively.

\begin{table}[htbp]
\renewcommand{\arraystretch}{1.3}
\caption{\ac{UKF} tolerances: Maximum bias and random error standard deviation.}
\label{tab: ukfvalues}
\centering
\begin{tabular}{lll}
\hline
\hline
Measurement       & Bias error                      & Random error \\
\hline
Roll rate, P      & $\pm9.98$ º/sec                 & $\pm8.51$ º/sec\\
Pitch rate, Q     & $\pm9.98$ º/sec                 & $\pm8.51$ º/sec\\
Yaw rate, R       & $\pm9.98$ º/sec                 & $\pm5.01$ º/sec\\
Accelerometers    & $\pm0.3$ m/s\textsuperscript{2} & $\pm0.7$ m/s\textsuperscript{2} \\
Magnetometers     & $\pm10.53$mG                    & $\pm21.73$mG  \\
\ac{GPS} Velocity & $\pm2.57$m/s                    & $\pm2.45$m/s\\
\hline
\hline
\end{tabular}
\end{table}

\begin{table}[htbp]
\renewcommand{\arraystretch}{1.3}
\caption{\ac{EKF} tolerances: Maximum bias and random error standard deviation.}
\label{tab: ekfvalues}
\centering
\begin{tabular}{lll}
\hline
\hline
Measurement       & Bias error                       & Random error\\
\hline
Roll rate, P      & $\pm1.50$ º/sec                  & $\pm1.50$ º/sec\\
Pitch rate, Q     & $\pm1.50$ º/sec                  & $\pm1.50$ º/sec\\
Yaw rate, R       & $\pm0.50$ º/sec                  & $\pm1.50$ º/sec\\
Accelerometers    & $\pm0.14$ m/s\textsuperscript{2} & $\pm0.32$ m/s\textsuperscript{2} \\
Magnetometers     & $\pm5.58$mG                      & $\pm11.43$mG  \\
\ac{GPS} Velocity & $\pm2.17$m/s                     & $\pm1.45$m/s\\
\hline
\hline
\end{tabular}
\end{table}

It can be noted how both algorithms are more sensitive to errors in $R$ than they are to errors in
$P$ and $Q$. This is because the information in the yaw angle is only provided by one of the
sensors, the magnetometer. While the information in pitch and roll angles are provided by the
accelerometer and the magnetometer.

In general, \ac{EKF} is more sensitive to biases and noises than the \ac{UKF}. In particular it is
very sensitive to the bias in $R$. Hence, the solution based on the \ac{UKF} is preferred.

\subsection{Simulation using real error values}

For this simulation, the real biases and noise magnitudes were extracted from our \ac{IMU}'s
datasheet. Table~\ref{tab: mems_err} shows them. It is interesting to compare these values to those
from tables \ref{tab: ukfvalues} and \ref{tab: ekfvalues}. It can be seen how some errors are
outside of the \ac{EKF} tolerances. So \ac{EKF} can not be used in our case.

\begin{table}
\renewcommand{\arraystretch}{1.3}
\caption{Typical \ac{MEMS} bias and random error standard deviation.}
\label{tab: mems_err}
\centering
\begin{tabular}{lll}
\hline
\hline
Measurement       & Bias error                       & Random error\\
\hline
Roll rate, P      & $\pm3.00$ º/sec                  & $\pm1.00$ º/sec\\
Pitch rate, Q     & $\pm3.00$ º/sec                  & $\pm1.00$ º/sec\\
Yaw rate, R       & $\pm3.00$ º/sec                  & $\pm1.00$ º/sec\\
Accelerometers    & $\pm0.05$ m/s\textsuperscript{2} & $\pm0.009$ m/s\textsuperscript{2}\\
Magnetometers     & $\pm4.00$mG                      & $\pm1.25$mG\\
\ac{GPS} Velocity & $\pm0.5$m/s                      & $\pm1.5$m/s\\
\hline
\hline
\end{tabular}
\end{table}

\begin{figure}[htbp]
\centering
\includegraphics[width=1\linewidth]{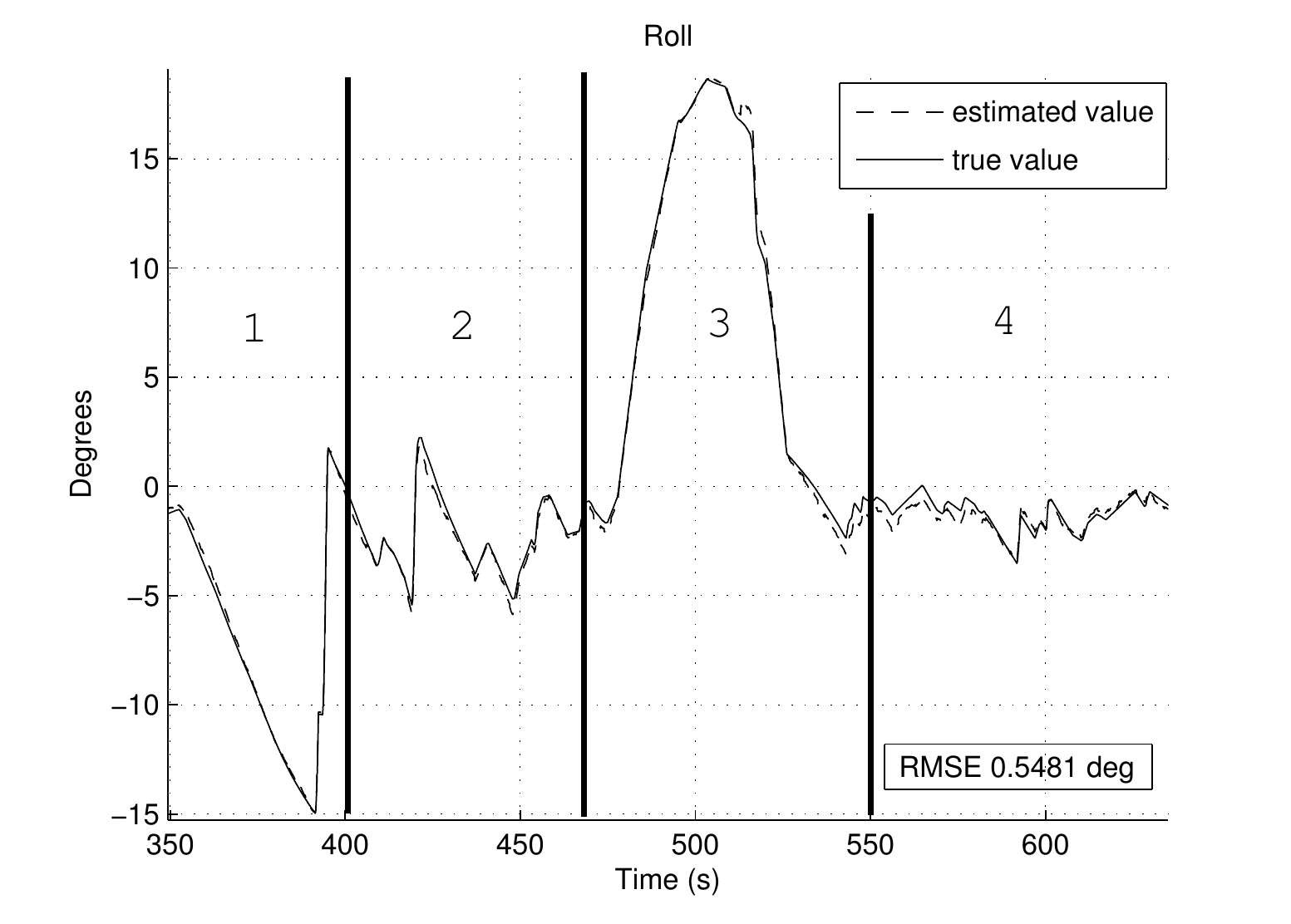}
\caption{Roll angle simulated with \ac{MEMS} parameters.}
\label{fig: ukfroll}
\end{figure}

\begin{figure}[htbp]
\centering
\includegraphics[width=1\linewidth]{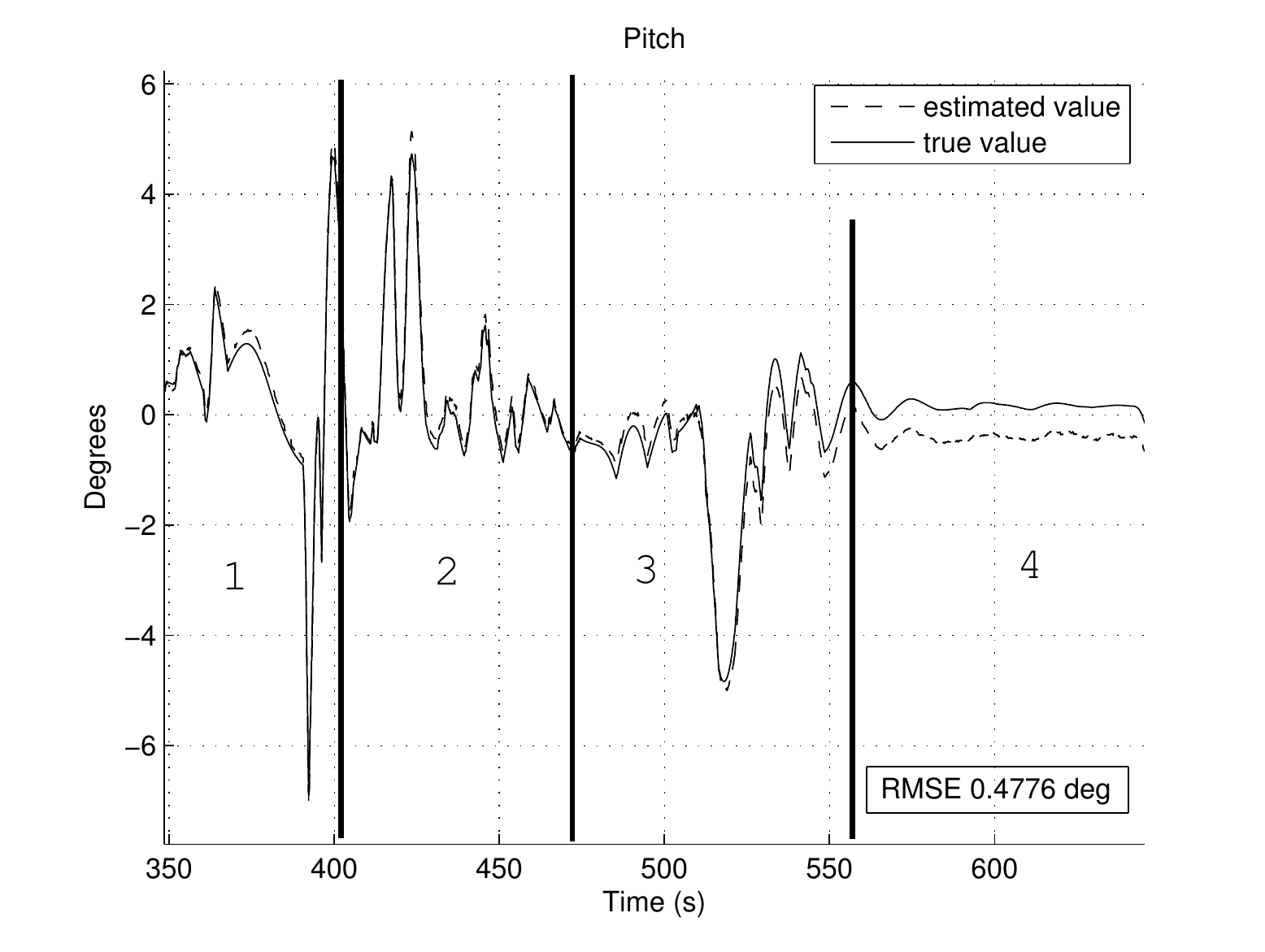}
\caption{Pitch angle simulated with \ac{MEMS} parameters.}
\label{fig: ukfpitch}
\end{figure}

\begin{figure}[htbp]
\centering
\includegraphics[width=1\linewidth]{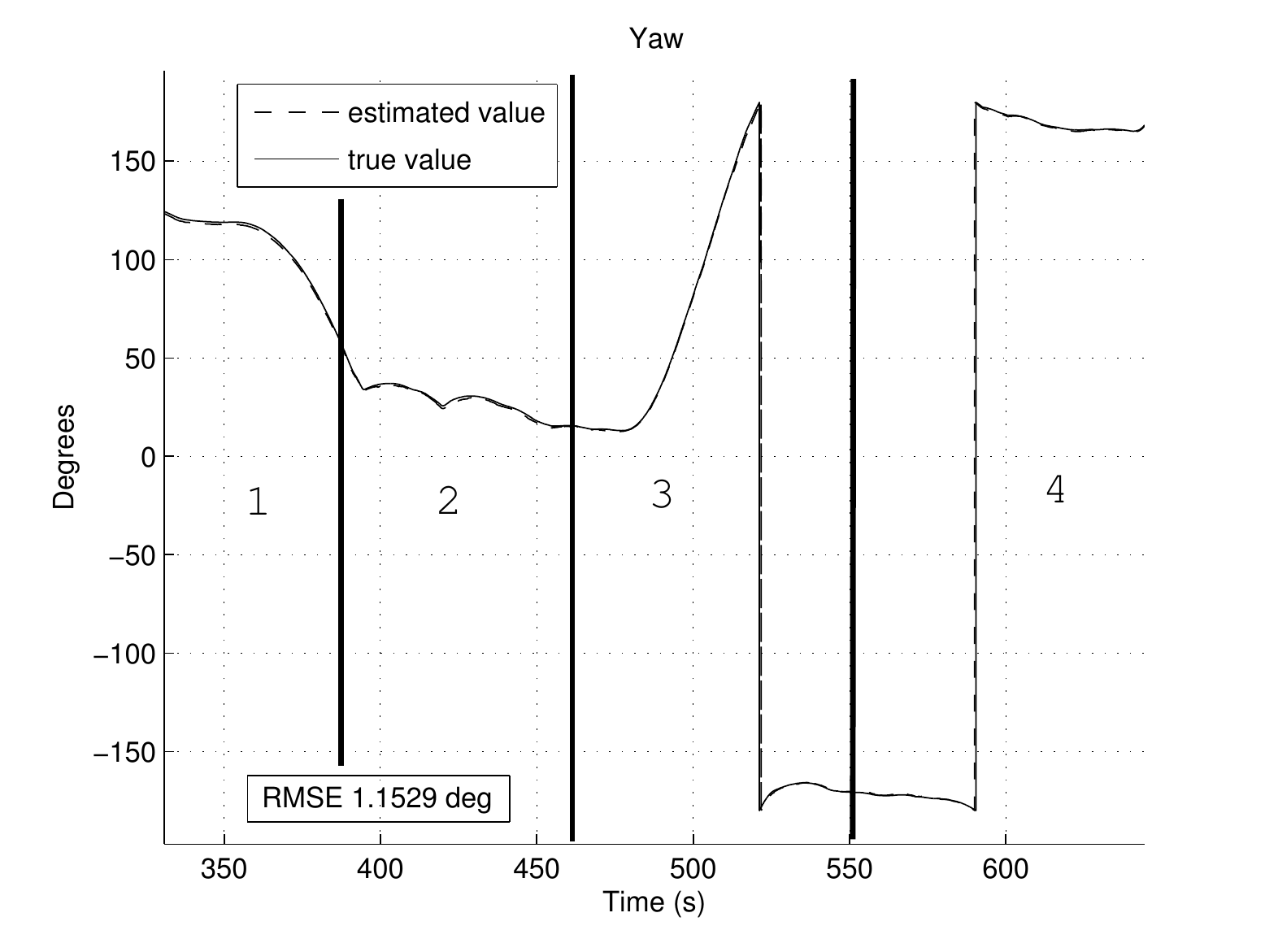}
\caption{Yaw angle simulated with \ac{MEMS} parameters.}
\label{fig: ukfyaw}
\end{figure}

Figures \ref{fig: ukfroll}, \ref{fig: ukfpitch} and \ref{fig: ukfyaw} show the behavior of the
\ac{UKF} algorithm in a typical flight in a windy scenario. Solid lines depict simulated angles
while the dashed ones depict estimations from the \ac{AHRS}. It should be noted that in
figure~\ref{fig: ukfyaw}, the abrupt changes near seconds 520 and 590 is due to the representation
of angles in the range $\pm 180º$. The agreement between the simulated and estimated values is
satisfactory.

Area marked with '1' shows a coordinated turn to the left, note how the yaw angle decreases. This
turn leads to the next area, marked with '2', which is a steady flight. The area marked with '3' is
a new coordinated turn to the right, note how the yaw angle increases. The final area, marked with
'4', is a new steady flight; the three angles remain mostly constant. It can be noted that the pitch
angle estimation in area '4' has a slight bias (of less than 0.3º), this is due to the magnetometers
biases added during the data corruption. These biases are not compensated through the algorithm
because of their negligible effect compared to the gyrometers biases. The pitch and roll angles
experience vibrations due to different gusts of wind.

This section has shown how the simulation results give confidence on the \ac{AHRS} using the
\ac{UKF}. The system can be used in field experiments. The following section covers field
experiments and the validation of the \ac{AHRS}.

\section{Field experiment results}

The target of the research is to use the \ac{AHRS} to feed a closed-loop controller in field
experiments. Aside from the simulation results presented in an earlier section, an experimental
validation of the system has been done.

\begin{figure}[htbp]
\centering
\includegraphics[width=1\linewidth]{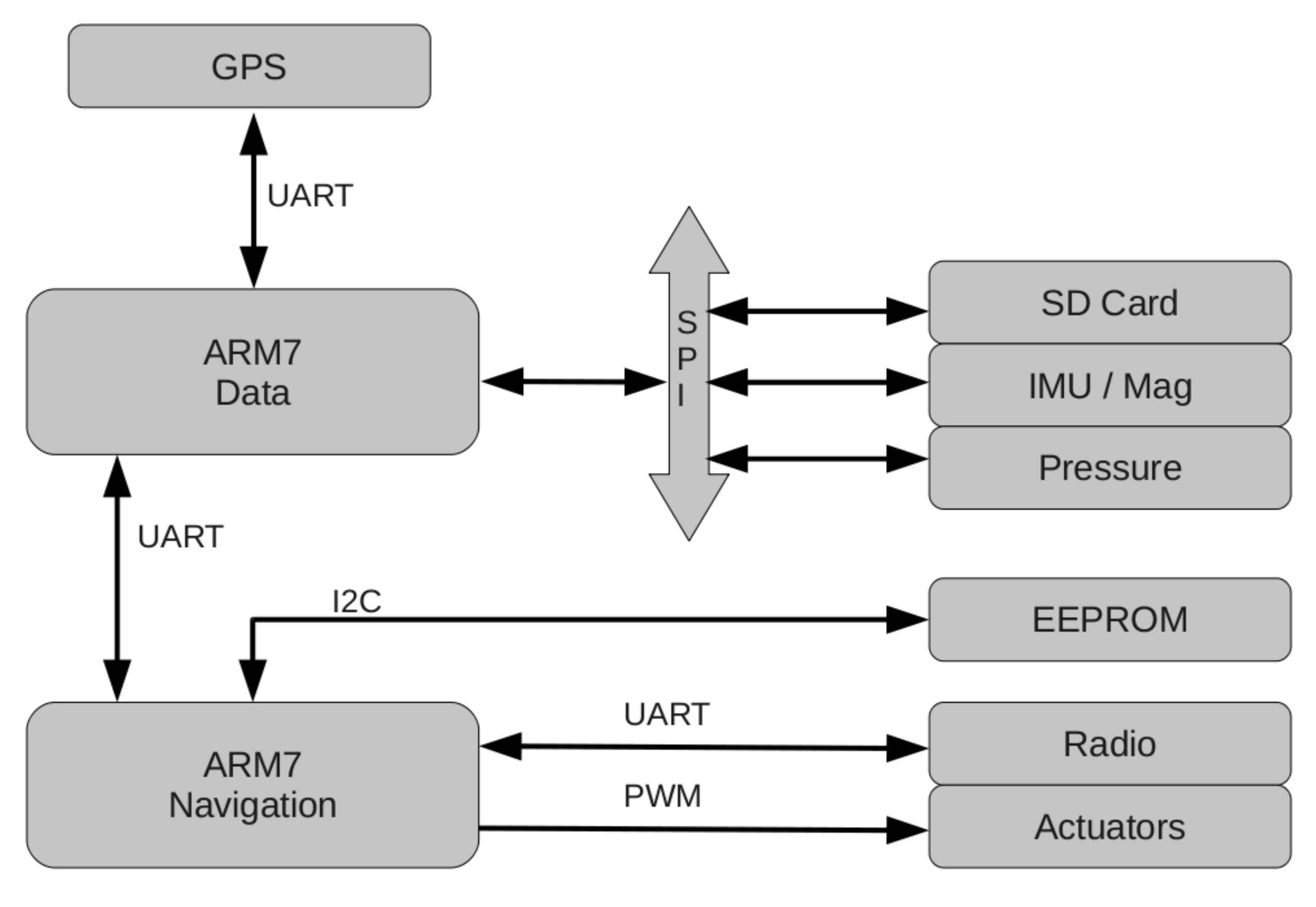}
\caption{Functional diagram of the autopilot.}
\label{fig: schhw}
\end{figure}

Experiments were carried out using a small fixed-wing \ac{UAV}. An on-board autopilot
hardware has been designed and built for this \ac{UAV}. Figure~\ref{fig: schhw} shows a functional
diagram of the autopilot.

Two ARM7 microcontrollers are used for sensor data handling and navigation algorithms. Both are
connected with a UART channel. Flight data are measured by different sensors: a \ac{GPS} receiver,
an \ac{IMU} with magnetometers, and four pressure sensors. One of the pressure sensors is used as a
barometer for altitude measurement and the rest, one per axis, are connected to Pitot tubes for
air-speed measurement. Sensor data are stored in an SD card for experiment analysis.  The \ac{IMU}
used is the ADIS16405 from Analog Devices.

\begin{figure}[htbp]
\centering
\includegraphics[width=1\linewidth]{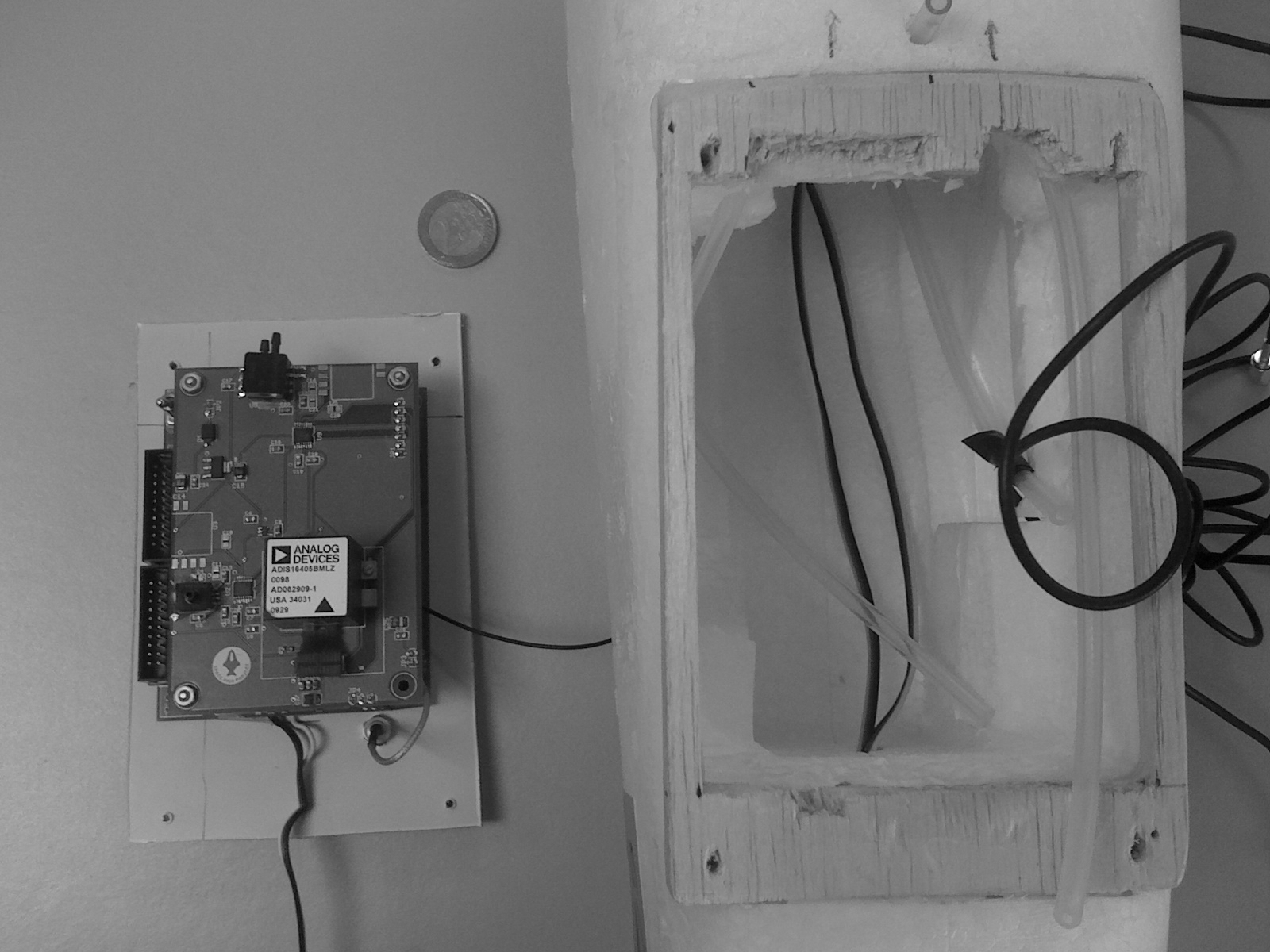}
\caption{Photograph of the on-board hardware next to the \ac{UAV}.}
\label{fig: hw}
\end{figure}

The navigation microcontroller receives processed sensor data from the other microcontroller, and
sends this data over the radio link. In future, when the control loop is closed, this
microcontroller will send the PWM signals to the actuators. Figure~\ref{fig: hw} shows a picture of
the autopilot.

The target of the experiments is to assess the accuracy of the estimation of each of the three
attitude angles. Therefore, experiments are designed to excite the \emph{single modes} of the
system, that is, pure roll, pure pitch and pure yaw. However, due to the nature of the system, it is
not possible to do so in practice; there is always some coupling between motions. We tried to excite
these modes during the experiments. In addition, typical maneuvers such as coordinated turns were
also included. A ground station was built to receive data from the \ac{UAV} using a radio link. The
flight is manually controlled using a conventional radio control unit.

For validation purposes, data coming from independent sensors (not used by the estimation algorithm)
have been considered. For the roll angle, a computer vision system is used. The GPS velocity is used
to validate the yaw angle.

\begin{figure}[htbp]
\centering
\includegraphics[width=1\linewidth]{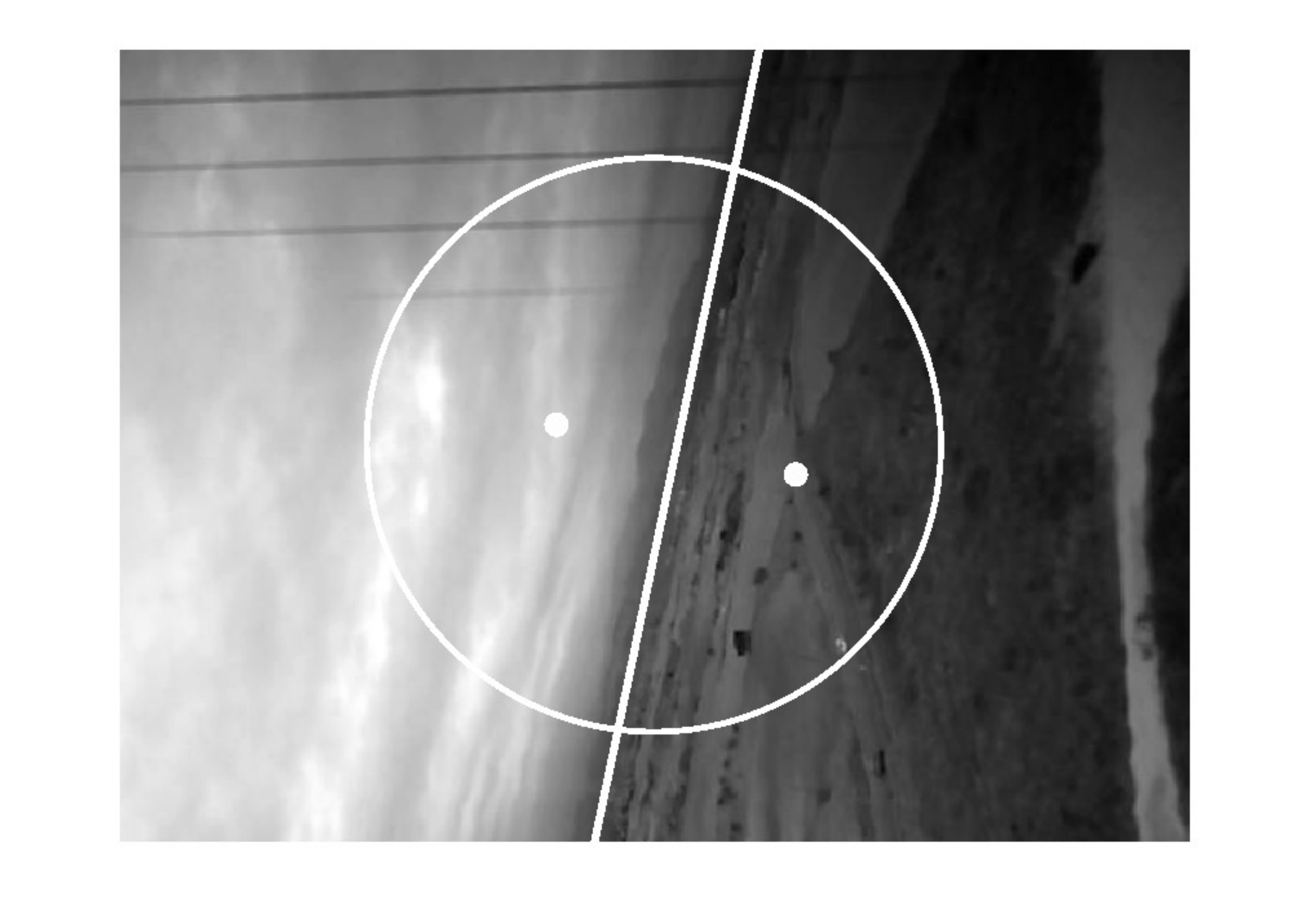}
\caption{One of the frames processed by the vision system.}
\label{fig: visionroll}
\end{figure}

The vision system uses a small camera attached to the \ac{UAV}. An algorithm was developed to obtain
the roll angle from the video measuring the slope of the horizon. The algorithm is based on
\cite{Bazin2008} \cite{Yuan2010} \cite{Pereira2008}. Figure~\ref{fig: visionroll} shows one of the
frames taken during the flight and processed by the vision system. The results of this system have
an uncertainty of $\pm 3º$.

\begin{figure}[htbp]
\centering
\hspace*{-0.05\linewidth}
\includegraphics[width=1.1\linewidth]{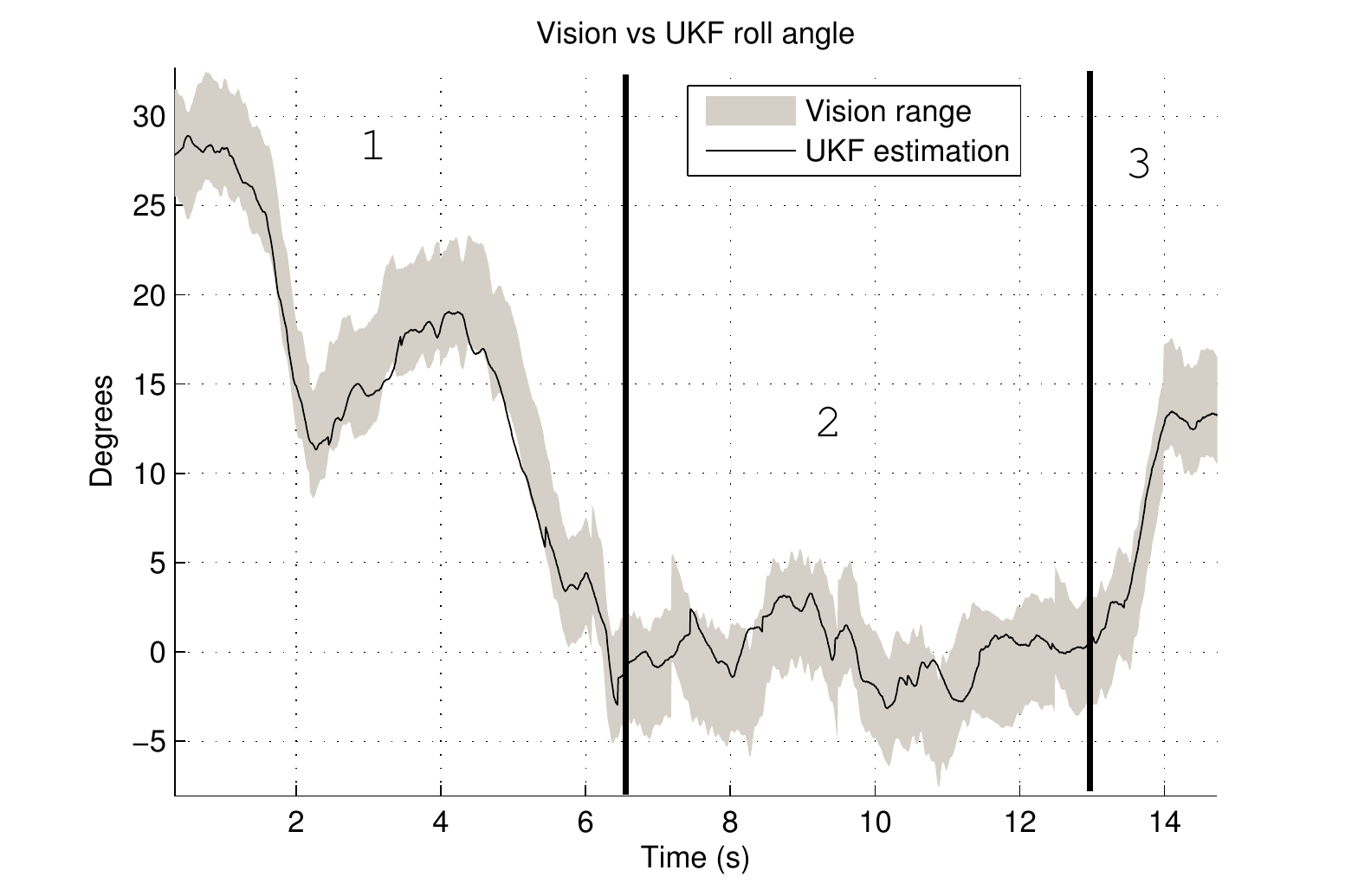}
\caption{Roll angle comparison between vision system and \ac{UKF} estimation.}
\label{fig: visionukfroll}
\end{figure}

Figure~\ref{fig: visionukfroll} shows a comparison between the estimated roll angle and the
measurements from the vision system. The area marked with '1' depicts the end of a turn to the
right. Is is followed by a steady flight in the area marked with '2', finally leading to another
turn to the right in the area marked with '3'. The results of the comparison are clearly
satisfactory.

The actual experiment took place under adverse atmospheric conditions, in particular, there were
gusts of wind. So when there is strong crosswind, the yaw angle differs from the heading of
the plane. The heading is measured using the \ac{GPS} velocity with an uncertainty of $\pm 10º$. It
can be noted that the \ac{GPS} velocity is only used to subtract the centrifugal contribution of the
accelerometers (see equation~\eqref{eq: centrifugal}) in the estimation algorithm. Therefore, it can
be taken as an independent system for validation purposes.

\begin{figure}[htbp]
\centering
\hspace*{-0.05\linewidth}
\includegraphics[width=1.1\linewidth]{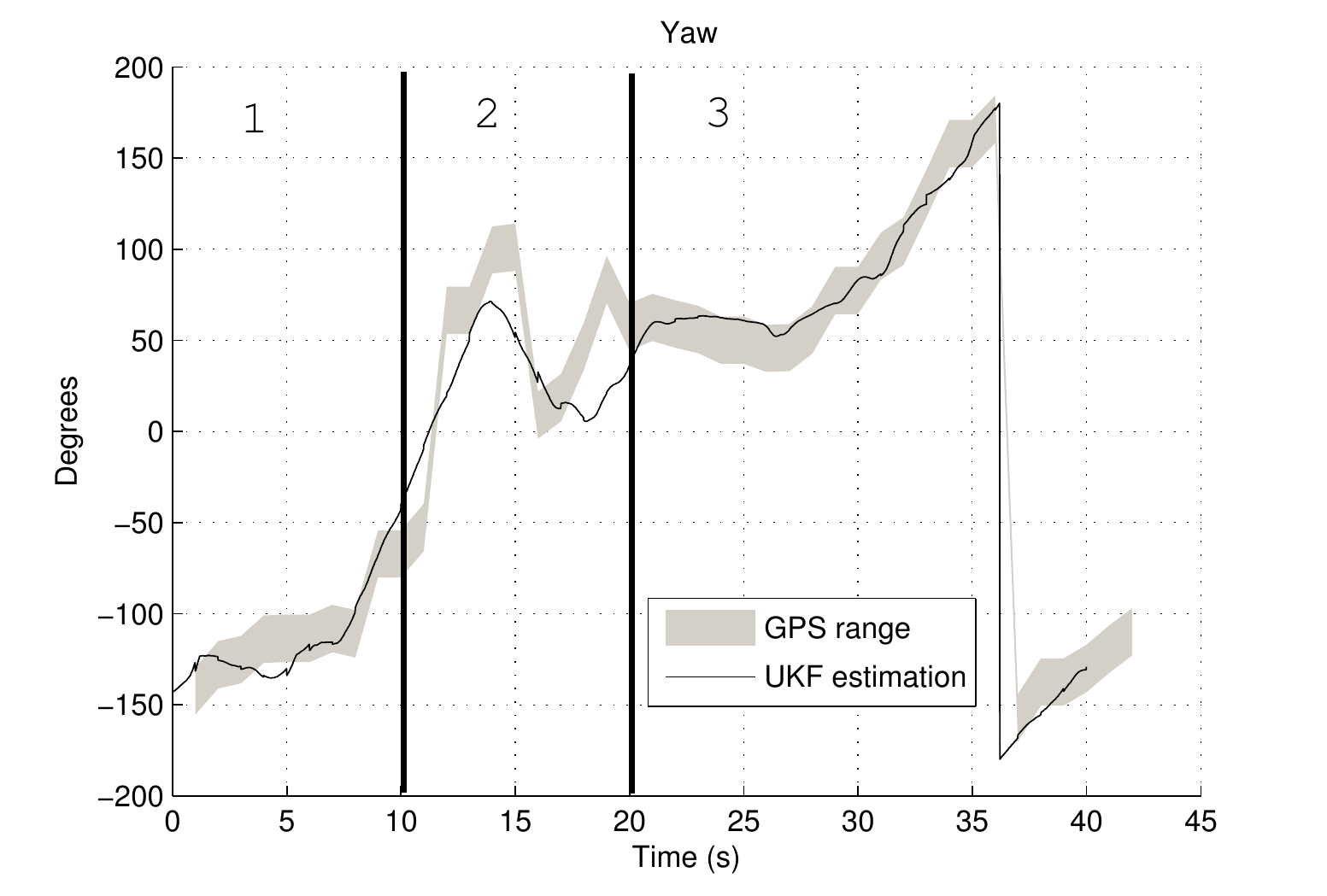}
\caption{Yaw angle comparison between \ac{GPS} system and \ac{UKF} estimation.}
\label{fig: gpsukfyaw}
\end{figure}

Figure~\ref{fig: gpsukfyaw} shows a comparison of the estimation of the yaw angle and the heading
measured by the \ac{GPS}. The figure corresponds to a turn of 360º. During the area marked with '1',
the \ac{UAV} faces the wind gusts. The area marked with '2' shows the effect of crosswind; it makes
the yaw angle and the heading diverge. In other words, the \ac{UAV} does not move in the direction
it points to. Finally, throughout the area marked with '3', the \ac{UAV} has tail wind and the yaw
angle closely follows the \ac{GPS} heading.

With the sensors we are currently using, it is not possible to have an independent measurement of
the pitch angle. However, due to the formulation of the algorithm using quaternions, the pitch angle
is strongly coupled to the roll and yaw angles. Therefore, it can be assumed that if both the roll
and yaw angles are correctly estimated, the pitch angle is correctly estimated too.

\section{Conclusions.}

This paper considered the problem of attitude estimation of an \ac{UAV} in order to establish
closed-loop control. The mathematical formulation of the problem has been presented in terms of
quaternions. The on-board \ac{AHRS} is based on a \ac{MEMS} \ac{IMU}.

A widely used estimator is the Kalman Filter. The kinematic model of aircraft attitude is highly
non-linear, so a version of the Kalman filter able to cope with non-linearities is needed. Two of
these versions, have been studied: the \ac{EKF} and the \ac{UKF}. A common solution in the satellite
attitude estimation practice is the \ac{TRIAD} algorithm and it has been used as the observation
model in the \ac{UKF} framework in this paper.

Previous studies in the attitude determination field give the same confidence to the sensors
throughout the whole \ac{UAV} mission. This is unsuitable for missions that mix both acrobatic and
non-acrobatic maneuvers. Using the \ac{TRIAD} algorithm it is easy to select the most reliable
sensors along the different phases of a flight. Criteria for this selection have been described.

A simulation framework based on XPlane 9 has been introduced. This simulation corresponds to the
characteristics of the experimental \ac{UAV}. Using simulation results, it was found that the
\ac{UKF} shows better performance than the \ac{EKF}. Therefore, the \ac{UKF} has been used in the
final version of the \ac{AHRS}.

The performance of the algorithm has been assessed using field experiments. Using independent
sensors it has been checked that the estimation algorithm gives good results. A possible idea for
future research is to include the information from these sensors into the \ac{AHRS} itself. Since
the estimation results are encouraging the next experimental work will be to feed the estimations to
a closed-loop controller.

Due to space constraints only some snapshots of the experimental results have been shown. However,
the complete set of data confirm the good quality of the estimations. This gives confidence on the
algorithm, which is easy to implement and it can be run on an on-board microcontroller.

\useRomanappendicesfalse
%




%

%

%

\bibliographystyle{IEEEtran.bst}
\bibliography{IEEEabrv,modif_refs}

\begin{thebibliography}{10}
\providecommand{\url}[1]{#1}
\csname url@rmstyle\endcsname
\providecommand{\newblock}{\relax}
\providecommand{\bibinfo}[2]{#2}
\providecommand\BIBentrySTDinterwordspacing{\spaceskip=0pt\relax}
\providecommand\BIBentryALTinterwordstretchfactor{4}
\providecommand\BIBentryALTinterwordspacing{\spaceskip=\fontdimen2\font plus
\BIBentryALTinterwordstretchfactor\fontdimen3\font minus
  \fontdimen4\font\relax}
\providecommand\BIBforeignlanguage[2]{{%
\expandafter\ifx\csname l@#1\endcsname\relax
\typeout{** WARNING: IEEEtran.bst: No hyphenation pattern has been}%
\typeout{** loaded for the language `#1'. Using the pattern for}%
\typeout{** the default language instead.}%
\else
\language=\csname l@#1\endcsname
\fi
#2}}

\bibitem{Maza2010}
I.~Maza, F.~Caballero, J.~Capitan, J.~M. de~Dios, and A.~Ollero, ``Firemen
  monitoring with multiple uavs for search and rescue missions,'' in
  \emph{Proc. IEEE Workshop on Security, Safety and Rescue Robotics (SSRR)},
  2010.

\bibitem{Cole2006}
D.~T. Cole, S.~Sukkaireh, and A.~H. Göktogan, ``System development and
  demonstration of a uav control architecture for information gathering
  missions,'' \emph{Journal of Field Robotics}, no.~26, pp. 417--440, 2006.

\bibitem{Xia2011}
Y.~Xia, Z.~Zhu, M.~Fu, and S.~Wang, ``Attitude tracking of rigid spacecraft
  with bounded disturbances,'' \emph{IEEE Transactions on Industrial
  Electronics}, vol.~58, no.~2, pp. 647--659, 2011.

\bibitem{Zheng2011}
B.~Zheng and Y.~Zhong, ``Robust attitude regulation of a 3-dof helicopter
  benchmark: Theory and experiments,'' \emph{IEEE Transactions on Industrial
  Electronics}, vol.~58, no.~2, pp. 660--670, 2011.

\bibitem{Zheng2011b}
Z.~Zhu, Y.~Xia, and M.~Fu, ``Adaptive sliding mode control for attitude
  stabilization with actuator saturation,'' \emph{IEEE Transactions on
  Industrial Electronics}, vol.~PP, 2011.

\bibitem{Cai2008}
G.~Cai, B.~Chen, K.~Peng, M.~Dong, and T.~Lee, ``Modeling and control of the
  yaw channel of a uav helicopter,'' \emph{IEEE Transactions on Industrial
  Electronics}, vol.~55, no.~9, pp. 3426--3434, 2008.

\bibitem{Spinka2011}
O.~\v{S}pinka, O.~Holub, and Z.~Hanzálek, ``Low-cost reconfigurable control
  system for small uavs,'' \emph{IEEE Transactions on Industrial Electronics},
  vol.~58, no.~3, pp. 880--889, 2011.

\bibitem{nasa97}
L.~B. Stephen A.~Whitmore, Mike~Fife, ``Development of a closed-loop strap down
  attitude system for an ultrahigh altitude flight experiment,'' NASA Technical
  Memorandum 4775, 1997.

\bibitem{Julier1997}
S.~J. Julier and J.~K. Uhlmann, ``A new extension of the kalman filter to
  nonlinear systems,'' in \emph{Proc. Int. Symp. Aerospace/Defense Sensing,
  Simul. and Controls, Orlando, FL}, 1997, pp. 182--193.

\bibitem{vanMerwe}
E.~Wan and R.~van~der Merwe, ``The unscencted kalman filter,'' in \emph{Kalman
  Filtering and Neural Networks}, S.~Haykin, Ed.\hskip 1em plus 0.5em minus
  0.4em\relax Wiley, 2001, ch.~7.

\bibitem{Julier2004}
S.~J. Julier and J.~K. Uhlmann, ``Unscented filtering and nonlinear
  estimation,'' in \emph{Proceedings of the IEEE}, vol.~92, no.~3, 2004, pp.
  401--422.

\bibitem{Jaz1970}
A.~H. Jazwinski, \emph{Stochastic Processes and Filtering Theory}.\hskip 1em
  plus 0.5em minus 0.4em\relax Academic Press, Inc., 1970.

\bibitem{Reif1999}
K.~Reif, S.~Gunther, E.~Yaz, and R.~Unbehauen, ``Stochastic stability of the
  discrete-time extended kalman filter,'' \emph{IEEE Transactions on Automatic
  Control}, vol.~44, no.~4, pp. 714--728, April 1999.

\bibitem{Gustaf2002}
F.~Gustafsson, F.~Gunnarsson, N.~Bergman, U.~Forssell, J.~Jansson, R.~Karlsson,
  and P.-J. Nordlund, ``Particle filters for positioning, navigation, and
  tracking,'' \emph{IEEE Transactions on Signal Processing}, vol.~50, no.~2,
  pp. 425--437, Feb. 2002.

\bibitem{Won2010}
S.~P. Won, W.~W. Melek, and F.~Golnaraghi, ``A kalman/particle filter-based
  position and orientation estimation method using a position sensor/inertial
  measurement unit hybrid system,'' \emph{IEEE Transactions on Industrial
  Electronics}, vol.~57, no.~5, pp. 1787--1798, May 2010.

\bibitem{Teixeira2010}
B.~O. Teixeira, L.~A. Tôrres, P.~Iscold, and L.~A. Aguirre, ``Flight path
  reconstruction - a comparison of nonlinear kalman filter and smoother
  algorithms,'' \emph{Aerospace Science and Technology}, vol. In Press,
  Corrected Proof, 2010.

\bibitem{Marins2001}
J.~Marins, X.~Yun, E.~Bachmann, R.~McGhee, and M.~Zyda, ``An extended kalman
  filter for quaternion-based orientation estimation using marg sensors,'' in
  \emph{Proc. IEEE/RSJ International Conference on Intelligent Robots and
  Systems.}, vol.~4, 2001, pp. 2003--2011.

\bibitem{Hale2004}
M.~J. Hale, P.~Vergez, M.~J. Meerman, and Y.~Hashida, ``Kalman filtering and
  the attitude determination and control task,'' in \emph{Proc. AIAA Space
  Conference and Exhibit}, Sept. 2004, paper AIAA-2004-6018.

\bibitem{Qi2008}
J.~Qi, J.~Han, and Z.~Wu, ``Rotorcraft uav actuator failure estimation with
  kf-based adaptive ukf algorithm,'' in \emph{American Control Conference},
  June 2008, pp. 1618--1623.

\bibitem{Hasan2009}
A.~M. Hasan, K.~Samsudin, A.~R. Ramli, R.~S. Azmir, and S.~A. Ismaeel, ``A
  review of navigation systems (integration and algorithms),'' \emph{Australian
  Journal of Basic and Applied Sciences}, vol.~3, no.~2, pp. 943--959, 2009.

\bibitem{Shuster1981}
M.~D. Shuster and S.~D. Oh, ``Three axis attitude determination from vector
  observations,'' \emph{Journal of Guidance and Control}, vol.~4, no.~1, pp.
  70--77, Jan. 1981.

\bibitem{Won2009}
S.~P. Won, F.~Golnaraghi, and W.~W. Melek, ``A fastening tool tracking system
  using an imu and a position sensor with kalman filters and a fuzzy expert
  system,'' \emph{IEEE Transactions on Industrial Electronics}, vol.~56, no.~5,
  pp. 1782--1792, May 2009.

\bibitem{Du2010}
D.~Du, L.~Liu, and X.~Du, ``A low-cost attitude estimation system for uav
  application,'' in \emph{Chinese Control and Decision Conference (CCDC)}, May
  2010, pp. 4489--4492.

\bibitem{Song2010}
S.~Lei, H.~Chang-qiang, W.~Xing-wei, and W.~Wen-chao, ``Research on a new
  eight-accelerometer configuration for attitude angle calculation of uav,'' in
  \emph{Chinese Control and Decision Conference (CCDC)}, May 2010, pp.
  4119--4123.

\bibitem{Wu2008}
W.~YongLiang, W.~TianMiao, L.~JianHong, W.~ChaoLei, and Z.~Chen, ``Attitude
  estimation for small helicopter using extended kalman filter,'' in \emph{IEEE
  Conference on Robotics, Automation and Mechatronics}, Sept. 2008, pp.
  577--581.

\bibitem{Jang2007}
J.~S. Jang and D.~Liccardo, ``Small uav automation using mems,'' \emph{{IEEE}
  Aerosp. Electron. Syst. Mag}, vol.~22, no.~5, pp. 30--34, May 2007.

\bibitem{Gebre2000}
D.~Gebre-Egziabher, G.~Elkaim, J.~Powell, and B.~Parkinson, ``A gyro-free
  quaternion-based attitude determination system suitable for implementation
  using low cost sensors,'' in \emph{{IEEE} Position Location and Navigation
  Symposium}, 2000, pp. 185--192.

\bibitem{Hadri2009}
A.~E. Hadri and A.~Benallegue, ``Sliding mode observer to estimate both the
  attitude and the gyro-bias by using low-cost sensors,'' in \emph{IEEE/RSJ
  International Conference on Intelligent Robots and Systems (IROS)}, Oct.
  2009, pp. 2867--2872.

\bibitem{Vaganay1993}
J.~Vaganay, M.~Aldon, and A.~Fournier, ``Mobile robot attitude estimation by
  fusion of inertial data,'' in \emph{Proc. {IEEE} International Conference on
  Robotics and Automation}, vol.~1, May 1993, pp. 277--282.

\bibitem{Bazin2008}
J.-C. Bazin, I.~Kweon, C.~Demonceaux, and P.~Vasseur, ``Uav attitude estimation
  by vanishing points in catadioptric images,'' in \emph{IEEE International
  Conference on Robotics and Automation (ICRA)}, May 2008, pp. 2743--2749.

\bibitem{Yuan2010}
H.-Z. Yuan, X.-Q. Zhang, and Z.-L. Feng, ``Horizon detection in foggy aerial
  image,'' in \emph{International Conference on Image Analysis and Signal
  Processing (IASP)}, Apr. 2010, pp. 191--194.

\bibitem{Pereira2008}
G.~Pereira, P.~Iscold, and L.~Torres, ``Airplane attitude estimation using
  computer vision: simple method and actual experiments,'' \emph{Electronics
  Letters}, vol.~44, no.~22, pp. 1303--1304, Oct. 2008.

\end{thebibliography}

%








\end{document}